\documentclass[10pt,twocolumn,letterpaper]{article}

\usepackage[pagenumbers]{iccv} 
\usepackage{kotex}
\usepackage{mathrsfs}

\definecolor{iccvblue}{rgb}{0.21,0.49,0.74}
\usepackage[pagebackref,breaklinks,colorlinks,allcolors=iccvblue]{hyperref}

\def\paperID{6277} 
\def\confName{ICCV}
\def\confYear{2025}

\usepackage{multirow}
\title{Fine-Tuning Visual Autoregressive Models for Subject-Driven Generation}

\author{Jiwoo Chung, Sangeek Hyun, Hyunjun Kim, Eunseo Koh, MinKyu Lee, Jae-Pil Heo$^{\ast}$\\
Sungkyunkwan University\\
{\tt\small \{jiwoo.jg, hse1032, arithu3, rainniee999, 2minkyulee, jaepilheo\}@gmail.com}
}

\newcommand\blfootnote[1]{%
  \begingroup
  \renewcommand\thefootnote{}\footnote{#1}%
  \addtocounter{footnote}{-1}%
  \endgroup
}

\begin{document}
\twocolumn[{
\maketitle
\begin{center}
\captionsetup{type=figure}
\includegraphics[width=0.99\textwidth]{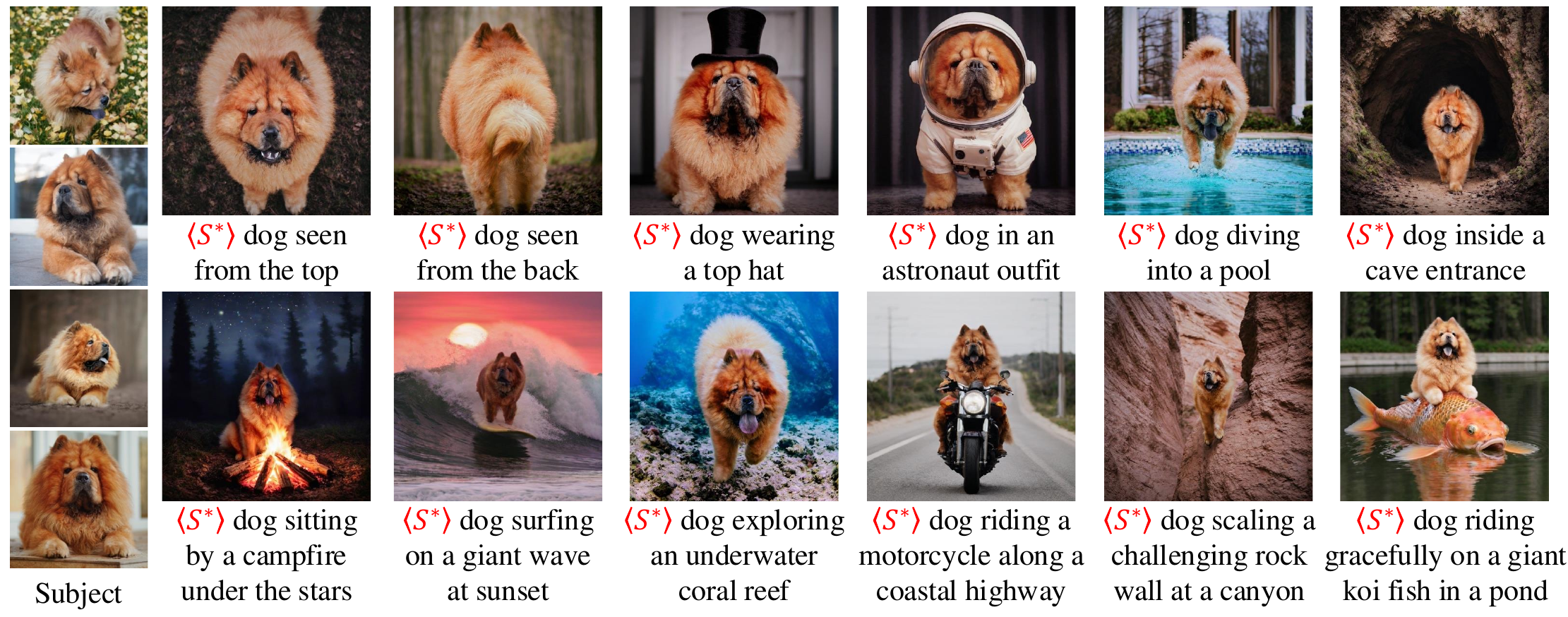}
\vspace{-0.2cm}
\captionof{figure}{
Our method achieves high subject fidelity and text alignment in subject-driven generation through Visual autoregressive models.
}
\vspace{0.2cm}
\end{center}
}]

\begin{abstract}
\blfootnote{
    $^\ast$ Corresponding author
}
Recent advances in text-to-image generative models have enabled numerous practical applications, including subject-driven generation, which fine-tunes pretrained models to capture subject semantics from only a few examples. While diffusion-based models produce high-quality images, their extensive denoising steps result in significant computational overhead, limiting real-world applicability.
Visual autoregressive~(VAR) models, which predict next-scale tokens rather than spatially adjacent ones, offer significantly faster inference suitable for practical deployment. In this paper, we propose the first VAR-based approach for subject-driven generation. However, na\"{\i}ve fine-tuning VAR leads to computational overhead, language drift, and reduced diversity. To address these challenges, we introduce selective layer tuning to reduce complexity and prior distillation to mitigate language drift.
Additionally, we found that the early stages have a greater influence on the generation of subject than the latter stages, which merely synthesize minor details. 
Based on this finding, we propose scale-wise weighted tuning, which prioritizes coarser resolutions for promoting the model to focus on the subject-relevant information instead of local details. 
Extensive experiments validate that our method significantly outperforms diffusion-based baselines across various metrics and demonstrates its practical usage.
Codes are available at \href{https:/github.com/jiwoogit/ARBooth}{github.com/jiwoogit/ARBooth}.

\end{abstract}    
\section{Introduction}

\label{sec:intro}
With recent advances in text-to-image (T2I) generative models~\cite{rombach2022high, hierarchical, imagereward, flux2024, peebles2023scalable}, numerous real-world applications~\cite{avrahami2022blended, ruiz2023dreambooth,p2p,chung2024style,Instinpaint,suppresseot, chai2023stablevideo, couairon2022diffedit, hertz2023prompt} have emerged and gained widespread adoption. Among these applications, subject-driven generation~(also called text-to-image personalization)~\cite{nam2024dreammatcher,galimage,ruiz2023dreambooth,dragdiffusion} represents the content of target subject using unique text embedding by optimizing the parameters or embeddings with only a few examples.

Most existing methods for subject-driven generation rely on pretrained diffusion models~\cite{rombach2022high, peebles2023scalable}, leveraging their ability to produce diverse and visually plausible images.
However, the diffusion process inherently requires a large number of denoising steps~(e.g., 50 DDIM steps)~\cite{songdenoising}, incurring significant computational cost.
This inefficiency limits the applicability of diffusion-based methods in real-world scenarios, where fast inference and computational efficiency is crucial.

Recently, a novel family of generative models called Visual autoregressive~(VAR)~\cite{tian2025visual} has been introduced.
Unlike a typical autoregressive model that first encode image into tokens and sequentially predicts tokens in a raster scan order, it reformulates autoregressive modeling to predict next-scale tokens at once, instead of different locations.
This approach significantly accelerates inference speeds—approximately 20 times faster than diffusion models—while achieving state-of-the-art performance in image quality, data efficiency, and scalability~\cite{tian2025visual}. Notably, recent work~\cite{han2024infinity} further demonstrated VAR's scalability in large-scale T2I generation tasks. Despite these promising advantages, the application of VAR models to subject-driven generation remains unexplored.

In this paper, we present the first investigation of subject-driven generation within the VAR framework. Drawing inspiration from diffusion-based approaches~\cite{kumari2023multi,ruiz2023dreambooth}, we initially explore fine-tuning both model parameters and textual embeddings to capture the target subject.
However, na\"{\i}ve fine-tuning presents challenges such as computational inefficiency, language drift, and reduced diversity, which have also been identified in language models~\cite{lu2020countering,lee2019countering} and diffusion-based approaches~\cite{kumari2023multi,ruiz2023dreambooth,fan2024dreambooth++}.

To effectively address these issues, we introduce selective layer tuning and prior distillation. Through our analysis, we observe that cross-attention and feed-forward layers in the VAR transformer exhibit larger weight updates.
As larger updates imply the significance of weights to capture the subject-relevant information~\cite{li2020few, kumari2023multi}, we selectively fine-tune the module mostly updated. 
This approach substantially reduces computational overhead without sacrificing performance.
Additionally, our proposed prior distillation leverages the pretrained model's semantic knowledge to produce diverse instances of the same subject class without requiring extensive external datasets. By utilizing the efficient inference of the VAR architecture, our approach enables end-to-end fine-tuning, effectively mitigating language drift and improving output diversity.

Furthermore, we analyze the generation impact of different scales in the VAR architecture.
Specifically, we observe that coarser-scale token maps have a greater influence on synthesizing the identity of subject, whereas finer-scale tokens have a relatively smaller impact.
Motivated by this observation, we propose scale-wise weighted tuning, which prioritizes coarser resolutions to effectively capture subject semantics while placing less emphasis on minor variations such as finer-scale details.
Extensive experiments demonstrate that our method accurately encodes the target concept while maintaining robust editing capabilities, significantly outperforming diffusion-based baselines.

Our main contributions are summarized as follows:
\begin{itemize}
    \item[--] 
    We introduce the first attempt for optimization-based subject-driven generation method that utilizes a large-scale pretrained VAR model.
    \item[--] 
    We propose three components based on our findings to mitigate several challenges; selective layer tuning, scale-wise weighted tuning, and prior distillation.
    \item[--] 
    Extensive experiments validate that our method significantly outperforms diffusion-based baselines in terms of subject fidelity, text alignment, and inference speed.
\end{itemize}

\section{Related Work}
\label{sec:relatedwork}

\noindent\textbf{Autoregressive visual generation.}
Autoregressive models~(AR) for visual generation typically follow a raster scan manner~\cite{chen2020generative, van2016conditional, reed2017parallel}. Early methods~\cite{van2017neural, esser2021taming, razavi2019generating, lee2022autoregressive, yuvector} employ vector quantization~(VQ) to transform image patches to discrete tokens, predicting the next token with transformer models. Building on this pipeline, numerous works~\cite{yu2022scaling, wang2024emu3, team2024chameleon, wang2024loong, kondratyuk2024videopoet} have scaled VQ-based transformers for advanced text-to-image and video generation tasks.
Recently, Visual autoregressive~(VAR)~\cite{tian2025visual} models have introduced a next-scale prediction framework, significantly improving image quality.
Infinity~\cite{han2024infinity} incorporates bit-wise quantization~\cite{yu2023language, zhao2024image} to enable large-scale text-to-image generation within this framework for scalability.
Our approach extends this paradigm by introducing VAR in subject-drive generation, effectively capturing diverse subjects through fine-tuning.

\noindent\textbf{Applications in AR.}
Recent developments such as Llamagen2~\cite{sun2024autoregressive} have enhanced autoregressive generation by improving image compression, scaling architectures, and improving training procedures.
Following this direction, ControlAR~\cite{li2024controlar} and EditAR~\cite{mu2025editar} propose methods for injecting spatial controls like edge maps or depth maps into autogressive models.
Despite VAR's demonstrated potential, few studies have explored its application. ControlVAR ~\cite{li2024controlvar} and Car~\cite{yao2024car} extend controllable generation to VAR frameworks using scale-wise prediction. However, subject-driven generation remains largely unexplored in VAR contexts. Our work addresses this gap by proposing novel fine-tuning strategies tailored for subject-driven tasks.

\begin{figure*}[t!]
    \includegraphics[width=1.0\textwidth]{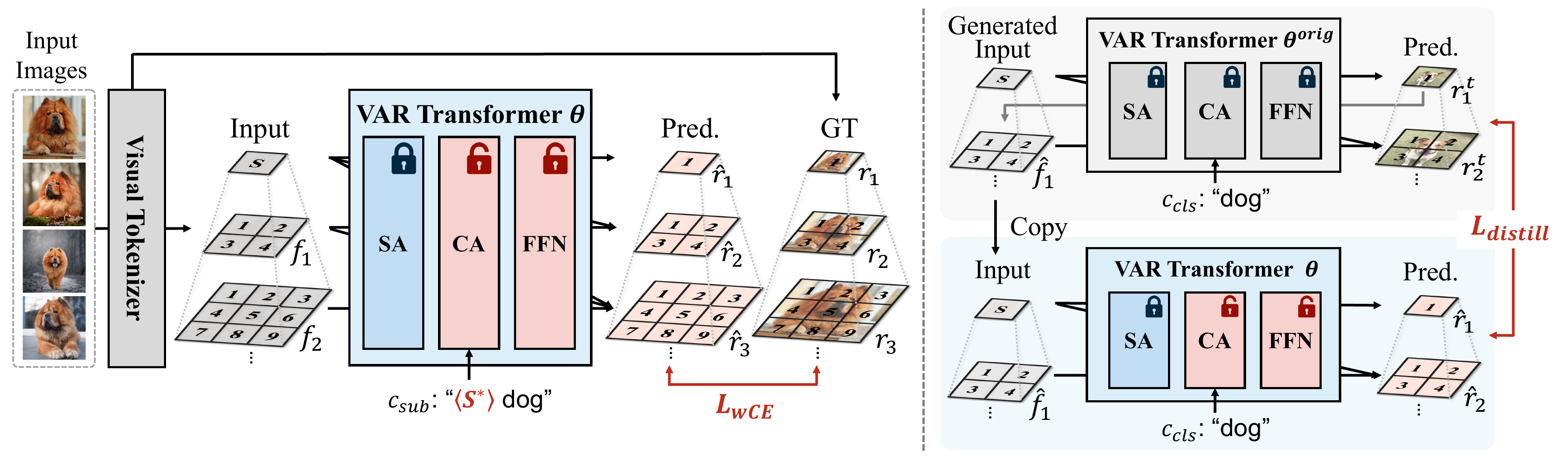}
    \centering
    \vspace{-0.4cm}
    \caption{\textbf{Overall Framework.}
    \textbf{(Left)~Subject-Driven Fine-Tuning.}
    We first encode the subject images into $K$ multi-scale token maps~($r_1, \dots, r_K$) using a visual tokenizer.
    Given a randomly sampled subject image, we fine-tune VAR transformer to reconstruct the corresponding token maps~($\hat{r}_1, \dots, \hat{r}_K$) to inject the concept of subject to the generator.
    During the optimization, we selectively fine-tune only the cross-attention~(CA) and feed-forward network~(FFN) layers~(Sec.~\ref{section:selective_layer_tuning}).
    Furthermore, we introduce a scale-wise weighted cross-entropy loss~($\mathcal{L}_\text{wCE}$) to efficiently focus coarser scales that contribute significantly to capturing subject semantics~(Sec.~\ref{section:scale_wise_weighted_tuning}).
    \textbf{(Right)~Prior Distillation.}
    To address language drift and enhance output diversity, we generate token maps using the original transformer~$\theta^\text{orig}$, conditioned on a class noun prompt~$c_{\text{cls}}$~(e.g., ``dog''). These token maps serve as guidance for the fine-tuned transformer, where a distillation-based loss~($\mathcal{L}_\text{distill}$) helps preserve the pretrained model's semantic prior~(Sec.~\ref{section:prior_distillation}).
    Finally, we jointly optimize the CA and FFN layers using both losses~($\mathcal{L}_\text{wCE},\mathcal{L}_\text{distill}$) to maintain subject fidelity and generative consistency.
     }
    \vspace{-0.0cm}
    \label{figure:main}
\end{figure*}

\noindent\textbf{Text-to-Image subject-driven generation.}
Early diffusion-based approaches represent subjects through optimized textual embeddings or full-model parameter fine-tuning, such as in Textual Inversion~\cite{galimage} and DreamBooth~\cite{ruiz2023dreambooth}.
Recent approaches~\cite{kumari2023multi, chen2023subject, jia2023taming, shi2024instantbooth, gal2023encoder, wei2023elite, xiao2024fastcomposer,li2023blip} emphasize efficient adaptations by either fine-tuning specific subsets of model parameters or introducing adapter modules; for example, CustomDiffusion~\cite{kumari2023multi} targets cross-attention layers. However, current methods predominantly rely on diffusion-based architectures. To our knowledge, we propose the first exploration of subject-driven text-to-image personalization in large-scale autoregressive models.

Moreover, our analysis also provide valuable insights for future VAR research. For example, we find that FFN layers in VAR are crucial, which aligns closely with recent observation in language models~\cite{geva2021transformer,geva2022transformer}. We expect that these insights can serve as a bridge for transferring techniques from language models to AR-based vision models.

\section{Preliminary}
\label{section:preliminary}

\subsection{Next-token Prediction}

Traditional autoregressive~(AR) models generate images by predicting tokens sequentially in a raster scan order~\cite{van2016conditional, van2016pixel,salimans2017pixelcnn++}. Specifically, a visual autoencoder fist encodes an image into a latent representation, which is then quantized into a sequence of discrete tokens~$x = ( x_1, x_2, \dots, x_T )$. To synthesize an image, the AR model iteratively predicts the probability distribution of the next token~$x_t$ given its preceding tokens~$x_{<t} = ( x_1, x_2, \dots, x_{t-1} )$ and condition $c$:
\begin{equation}
\label{eq:trad_ar_loss}
p(x) = \prod_{t=1}^T p(x_t \mid x_{<t}, c).
\end{equation}
After sampling tokens based on these predicted probabilities, the generated tokens are decoded into an image via the visual autoencoder.

\subsection{Next-scale Prediction}

Visual autoregressive~(VAR)~\cite{tian2025visual} models reformulate autoregressive prediction units from tokens at different spatial locations to tokens at different resolutions. That is, it predicts ``next-scale" tokens rather than sequential tokens.
Specifically, given an image~$I$, the visual encoder~$\mathcal{E}$ first encodes it into a feature map $f$. VAR then iteratively quantizes $f$ into $K$ multi-scale token maps $r = ( r_1, r_2, \dots, r_K )$.
Given previous token maps~$r_{<k}$ and a condition~$c$, the $k$-th scale tokens~$r_k$ are computed as:
\begin{equation}
\label{eq:ar_feat}
f_k = f - \sum_{i=1}^{k-1} \text{upsample}(\text{Lookup}(r_i)),
\end{equation}
\begin{equation}
\label{eq:ar_loss}
p(r) = \prod_{k=1}^K p(r_k \mid r_{<k}, c),
\end{equation}
where $\text{Lookup}(\cdot)$ is a lookup table mapping each token $r_i$ to its closest entry in a learnable codebook, and $\text{upsample}(\cdot)$ is a bilinear upsampling operation.

Recently, the next-scale prediction framework has demonstrated its effectiveness for large-scale text-to-image generation. Infinity~\cite{han2024infinity} adopts bit-wise quantization techniques~\cite{yu2023language, zhao2024image}, significantly expanding the available codeword space. As a result, it achieves text fidelity and aesthetic quality comparable to state-of-the-art diffusion-based methods, while providing faster inference speed. Thus, leveraging these advantages for subject-drive generation has the potential of extending this framework.

\begin{figure}[t!]
\includegraphics[width=0.9\columnwidth]{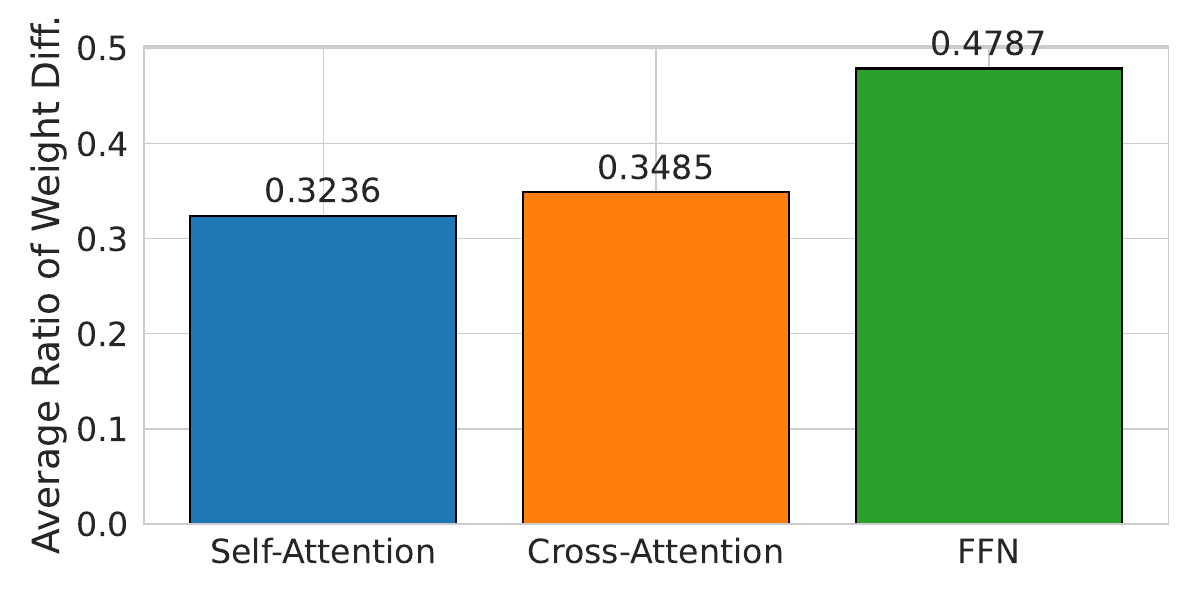}\centering
\vspace{-0.2cm}
    \caption{
    \textbf{Average Ratio of Weight Differences.}
    We measure the weight differences between the original model~($\theta^\text{orig}$) and its fine-tuned counterpart~($
    \theta$) using averaged ratios across layers.
    The averaged ratio is computed as $ |\theta^\text{orig} - \theta| / (|\theta^\text{orig}| + \epsilon)$ ,where $\epsilon$ is a small constant.
    Notably, the feed-forward networks~(FFN) exhibit significantly larger weight changes compared to other layers, indicating their importance to fine-tuning.
    }
    \label{figure:layer_weight_diff}
    \vspace{-0.1cm}
\end{figure}

\section{Method}
Given a set of $n$ subject images~$\mathcal{X}=\{ I_1, I_2, \dots, I_n \}$, existing subject-driven methods typically personalize text-to-image~(T2I) diffusion models or text embeddings~$c_\text{sub}$~(e.g., $\langle S^*\rangle$ dog) for represent specific subject semantics. During inference, these fine-tuned models generate images aligned with context prompts~(e.g., $\langle S^* \rangle$ dog in the desert).
Despite their successes, diffusion-based approaches are limited in practical applications thanks to slow inference speeds arising from extensive denoising steps. Recently, visual autoregressive~(VAR) models have been introduced as a promising solution due to their significantly faster inference capability.
We therefore aim to address subject-driven generation by harnessing the generative capabilities of a pretrained large-scale T2I VAR model.
However, na\"{\i}ve fine-tuning still brings several challenges, such as computational overhead, language drift, and reduced diversity. In this paper, we address these issues through novel techniques based on our comprehensive analysis.

\subsection{Selective Layer Tuning}
\label{section:selective_layer_tuning}
One straightforward approach for subject-driven generation is fine-tuning all layers of the generator with additional subject-specific text embeddings~\cite{ruiz2023dreambooth}.
However, excessive tuning of all layers increases computational cost and can degrade generated image quality due to overfitting, potentially harming the model's original generative capabilities.

To address this, we aim to identify a compact yet effective subset of parameters for fine-tuning, inspired by previous works~\cite{li2020few, kumari2023multi}. Specifically, we analyze layer-wise adaptation during personalization by measuring weight changes between the original and fully fine-tuned VAR models.
As reported in Fig.~\ref{figure:layer_weight_diff}, feed-forward layers exhibit the most significant weight changes. This suggests that these layers primarily integrate subject-specific information.
We first attempt to fine-tune only the feed-forward layers while keeping other layers fixed. However, we empirically observe a deterioration in text alignment, as shown in Fig.~\ref{figure:ablation_selective_layer}.
To ensure alignment with the target text and subject, we therefore selectively fine-tune only these two components, effectively capturing the subject concept while preserving overall generative capability.

\begin{figure*}
    \centering
    \begin{minipage}{0.65\textwidth} 
        \centering
        \includegraphics[width=1.0\linewidth]{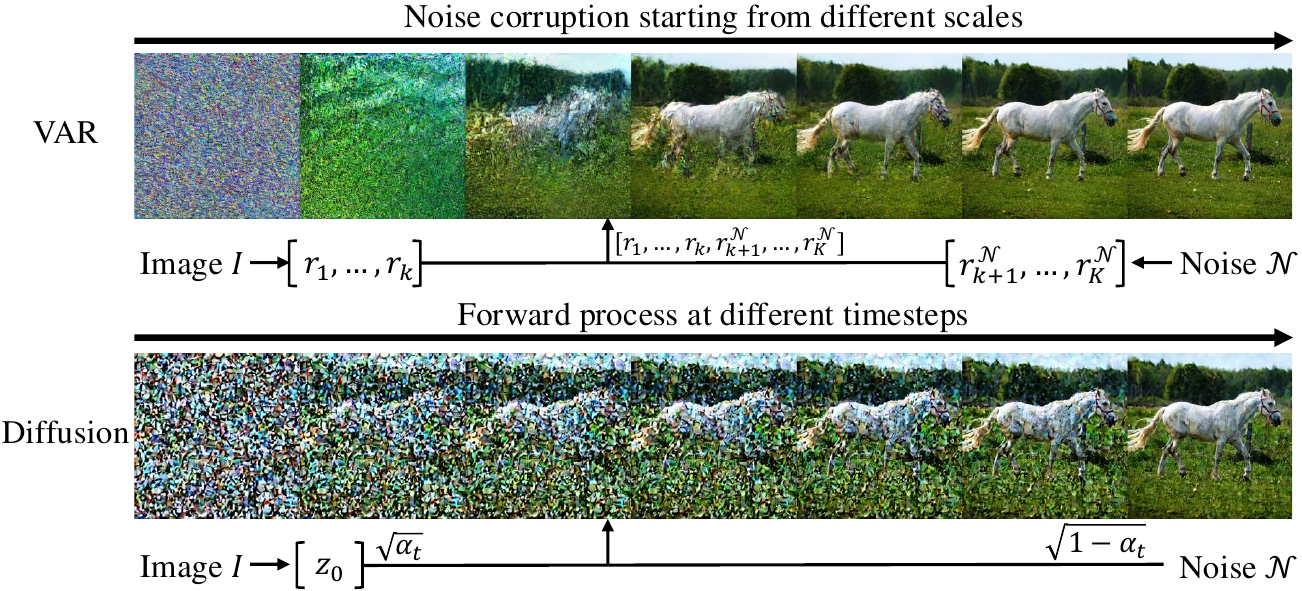}
        \vspace{-0.5cm}
        \caption{
        \textbf{Visualization of noise corruption across different stages.}
        (Top row)~Scale-wise noise corruption in VAR.
        To investigate each scale's contribution to image generation, we encode a given and noise image $I$ and $\mathcal{N}$ into token maps $r$ and $r^\mathcal{N}$, respectively.
        When synthesizing images by merging token maps from a given and noise images based on a specific scale, we observe coarser token maps contribute to the content while finer tokens adjust the local details.
        (Bottom row)~Timestep-based forward process in diffusion model. Similar to VAR, we encode the reference image into a latent representation~($z_0$)~\cite{rombach2022high} and visualize images at each timestep during the forward diffusion process~\cite{ho2020denoising, sohl2015deep}.
        Interestingly, scales of VAR and timesteps of diffusion exhibit similar disruption patterns, implying both coarser scale and higher time step contribute the actual content, while finer scale and lower timesteps adjust the details.
        Experimental details are available in Sec.~\ref{section:scale_wise_weighted_tuning}.
        }
        \label{fig:image_diff_plot}
    \end{minipage}
    \hfill
    \begin{minipage}{0.31\textwidth} 
        \centering
        \vspace{-0.0cm}
        \includegraphics[width=1.0\linewidth]{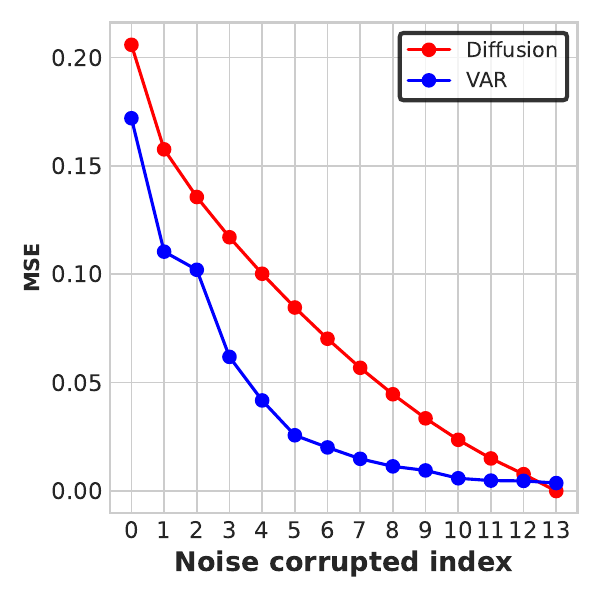}
        \vspace{-0.3cm}
        \caption{
        \textbf{MSE analysis between reference and noise-corrupted images.}
        We quantitatively analyze the impact of noise corruption at various scale and timesteps for VAR and diffusion models by measuring mean squared error~(MSE).
        Both architectures show similar trends, though VAR exhibits a stronger curvature, highlighting greater sensitivity at early stages.
        }
        \label{fig:image_diff_plot2}
    \end{minipage}
    \vspace{-0.1cm}
\end{figure*}
\subsection{Scale-wise Weighted Tuning}
\label{section:scale_wise_weighted_tuning}

Prior studies on diffusion models have revealed that their generative capability varies across different timesteps, with the distance between two timesteps in the diffusion process decreasing over timesteps~\cite{zheng2024non, zheng2024beta}.
Leveraging these observations, these methods prioritize training at earlier timesteps, where the models show significant updates, thus improving both efficiency and generation quality~\cite{zheng2024non, zheng2024beta}.

Similar to diffusion models, the VAR framework synthesizes images in a coarse-to-fine manner using a multi-scale residual quantization method. Consequently, its generative capability varies across different resolutions.
To specify this, we analyze the contribution of each scale in image generation. After identifying that each scale contributes differently to generation, we fine-tune the model at a scale-wise level to efficiently adapt to the subject concept.

Specifically, we first encode a given image $I$ and a noise image $\mathcal{N}$ into multi-scale token maps ($r_1, \dots, r_K$) and ($r^\mathcal{N}_1, \dots, r^\mathcal{N}_K$), respectively.
We then progressively construct noise-corrupted token representations by substituting finer-scale tokens from the subject image with those from the noise image~($r_1, \dots, r_k, r_{k+1}^{\mathcal{N}}, \dots, r_K^{\mathcal{N}}$). These corrupted representations are decoded to analyze their impact on the generation process.
As shown in Fig.~\ref{fig:image_diff_plot}, we observe that coarser resolutions lead to significant changes of the generated image whereas finer resolutions induce only minor variations in VAR. This behavior aligns with diffusion models~\cite{zheng2024non, zheng2024beta}, where the most drastic variations occur in the earlier stages.
Quantitatively, we also measure the mean square error~(MSE) between the reference image and its noise-corrupted counterpart across scales, indicating that VAR framework relies even more heavily on early-stage representations compared to diffusion models, as shown in Fig~\ref{fig:image_diff_plot2}.

Inspired by this observation, we prioritize training on scales with greater contributions during subject-driven fine-tuning. In detail, we introduce a scale-wise weighted cross-entropy loss~($\mathcal{L}_{\text{wCE}}$) which assigns higher weights to coarse scales and lower weights to fine scales:
\begin{equation}
\label{eq:finetune}
\mathcal{L}_{\text{wCE}} = - \sum_{k=1}^{K} w_k\log p_\theta(r_k \mid r_{<k}, c_{\text{sub}}),
\end{equation}
where $r_k$ denotes the multi-scale token map derived from subject images, and $w_k$ represents the hyperparameter to weight for each scale, following the order $w_1 \geq w_2 \geq \dots \geq w_K$. 
This strategy enables the fine-tuned model to effectively capture subject-specific semantics, while preserving its original generative capabilities.

\subsection{Prior Distillation}
\label{section:prior_distillation}
We further introduce prior distillation to address language drift and reduced diversity. Language drift, previously observed in language model fine-tuning~\cite{lu2020countering,lee2019countering}, occurs when a model gradually loses its pretrained knowledge, becoming overly focused on fine-tuned tasks. Similarly, we find that the fine-tuned VAR model tends to forget how to generate diverse instances within the same subject class. For example, when prompted with a generic class noun~(e.g., “a dog”), the personalized model consistently generates images resembling the specific fine-tuned subject.
In previous diffusion-based methods such as DreamBooth~\cite{ruiz2023dreambooth}, language drift is mitigated by first collecting numerous same-class images and jointly fine-tuning them alongside subject-specific images. However, this approach requires substantial data collection, which is often costly and impractical.

In our autoregressive framework, we propose an end-to-end prior preserving approach called~\textit{prior distillation}. Unlike diffusion-based methods that necessitate data collection, our approach leverages the fast inference capability of VAR to generate diverse instances of the subject's class.
Specifically, the efficiency of VAR's inference allow our distillation method to mitigate language drift without the burden of gathering large same-class datasets.




Concretely, we designate the large-scale pre-trained VAR transformer as the teacher in distillation process, as visualized in Fig.~\ref{figure:main}. Given a class prompt~($c_\text{cls}$), the original~(teacher) model~$\theta^\text{orig}$ generates multi-scale tokens. Instead of reconstructing images from these tokens and subsequently re-encoding them via the visual autoencoder and multi-scale quantizer, we directly utilize the original model's token outputs to guide predictions from the fine-tuned model. This approach effectively bypasses redundant encoding-decoding steps, significantly streamlining the distillation process.

The distillation objective is defined as follows:
\begin{equation}
\label{eq:distill}
\mathcal{L}_{\text{distill}} = \sum_{k=1}^{K} D_{\text{KL}}\Bigl(p_{\theta^t}(r^t_k \mid r^t_{<k}, c_{\text{cls}}) \,\Big\|\, p_{\theta}(r_k \mid r_{<k}, c_{\text{cls}})\Bigr),
\end{equation}
where $r^t_k$ denotes the token maps generated by the original model and $D_{\text{KL}}$ denotes that kl-divergence between two probabilities.
Through prior distillation, we effectively regulate the model’s fine-tuning by consistently exposing the original semantic knowledge. This ensures diverse generation and mitigates language drift.

In summary, we fine-tune the CA and FFN layers of the transformer and the text embedding while jointly optimizing the scale-wise weighted cross-entropy loss~($\mathcal{L}_\text{wCE}$) and prior distillation loss~($\mathcal{L}_\text{distill}$) to enhance subject fidelity and preserve generative consistency.
\section{Experiment}

\subsection{Implementation Details}
We conduct all experiments using Infinity-2B model~\cite{han2024infinity}, pretrained on the LAION~\cite{schuhmann2021laion}, COYO~\cite{kakaobrain2022coyo-700m}, and OpenImages~\cite{kuznetsova2020open} datasets. We follow the default configuration provided by Infinity~($K=13$) and apply classifier-free guidance~\cite{ho2021classifier} to logits as our default setting. The model is fine-tuned at a resolution of 1024 for 600 iterations with a batch size of 2 and use AdamW~\cite{loshchilovdecoupled} optimizer ($\beta_0=0.9, \beta_1=0.97$) with a learning rate of $6e-3$ on a single NVIDIA A6000 GPU. Additionally, we apply random resize and center crop augmentations following~\cite{von-platen-etal-2022-diffusers}.

\subsection{Evaluation Protocol}
\noindent\textbf{Dataset.}
We utilize the dataset from ViCo~\cite{ham2024personalized}, originally collected in previous works~\cite{ruiz2024hyperdreambooth, galimage, ruiz2023dreambooth}. This dataset comprises 16 subject concepts (5 living and 11 non-living), each paired with 31 textual prompts. For evaluation, we generate 8 images per concept-prompt pair, resulting in a total of 3,968 generated images.

\noindent\textbf{Baselines.}
We evaluate our proposed method against five diffusion-based state-of-the-art baselines: Textual Inversion~(TI)~\cite{galimage}, DreamBooth~(DB)~\cite{ruiz2023dreambooth}, CustomDiffusion~(CD)~\cite{kumari2023multi}, ELITE~\cite{wei2023elite}, ViCo~\cite{ham2024personalized}, and DreamMatcher~(DM)~\cite{nam2024dreammatcher}. For ELITE, we employ the official implementation. For TI, DB, CD, ViCo and DM, we adopt the pretrained weights or use the value provided by DM's publicly available implementation based on Stable Diffusion~(SD)~1.4~\cite{rombach2022high}, following their default configuration.

\noindent\textbf{Metrics.}
Following previous studies~\cite{galimage, ruiz2023dreambooth, nam2024dreammatcher, kumari2023multi}, we evaluate both subject fidelity and text prompt fidelity. For subject fidelity, we measure image similarity between the generated images and the reference subject using DINO~\cite{caron2021emerging} and CLIP~\cite{radford2021learning}, denoted as \textbf{I\textsubscript{dino}} and \textbf{I\textsubscript{clip}}, respectively. For text prompt fidelity, we compute CLIP image-text similarity by comparing the visual features of the generated images with the textual features of the prompts, excluding placeholders, and denote this as \textbf{T\textsubscript{clip}}.

For the ablation study, we measure a prior preservation metric~(PRES)~\cite{ruiz2023dreambooth}, defined as the average pairwise similarity between the DINO embeddings of generated images from random subjects of the prior class and real images of our target subject. A higher PRES value indicates that the random subjects resemble the target subject, suggesting that the model's ability to maintain the original has broken down. Furthermore, we compute diversity~(DIV)~\cite{ruiz2023dreambooth} using the average pairwise LPIPS~\cite{zhang2018unreasonable} distance between generated images of the same subject with the same prompt. These metrics capture a crucial property by indicating whether the model fails to preserve prior knowledge.

\subsection{Quantitative Comparison}
\begin{table}
    \centering
    \begin{tabular}{lccc|cc}
        \toprule
        \textbf{Model} & \textbf{I\textsubscript{dino}}~$\uparrow$ & \textbf{I\textsubscript{clip}}~$\uparrow$ & \textbf{T\textsubscript{clip}}~$\uparrow$ & \textbf{Time}~$\downarrow$ \\
        \midrule
        TI~\cite{galimage}     & 0.529          & 0.770          & 0.220 & 18s  \\
        DB~\cite{ruiz2023dreambooth}     & 0.640          & 0.815          & 0.236 & 18s  \\
        ELITE~\cite{wei2023elite}  & 0.584          & 0.783          & 0.223 & 11s  \\
        CD~\cite{kumari2023multi}     & 0.659          & 0.815          & 0.237 & 18s  \\
        ViCo~\cite{ham2024personalized}     & 0.643          & 0.816          & 0.228 & 15s  \\
        DM~\cite{nam2024dreammatcher}     & 0.682          & 0.823          & 0.234 & 32s  \\
        Ours   & \textbf{0.705} & \textbf{0.824} & \textbf{0.253} & \textbf{0.5s} \\
        \bottomrule
    \end{tabular}
    \vspace{-0.2cm}
    \caption{Quantitative comparison with baselines.}
    \vspace{-0.2cm}
    \label{table:model_comparison}
\end{table}

\begin{table}
    \centering
    \begin{tabular}{lccc}
        \toprule
        Preference & CD~\cite{kumari2023multi} & DM~\cite{nam2024dreammatcher} & Ours \\
        \midrule
        Subject Fidelity~$\uparrow$ & 15\% & 18\% & \textbf{67\%} \\
        Prompt Alignment~$\uparrow$ & 11\% & 10\% & \textbf{79\%} \\
        \bottomrule
    \end{tabular}
    \vspace{-0.2cm}
    \caption{Subject fidelity and prompt fidelity user preference.}
    \vspace{-0.4cm}
    \label{table:user_preference}
\end{table}

As shown in Tab.~\ref{table:model_comparison}, our method outperforms diffusion-based baselines on both \textbf{I\textsubscript{dino}} and \textbf{I\textsubscript{clip}}, indicating improved subject alignment. Notably, it also achieves better performance on \textbf{T\textsubscript{clip}}, reflecting enhanced alignment with the text prompt. Although these metrics often involve a trade-off, our method achieves higher values in both, demonstrating a strong enhancement.

In addition, we measure the inference time for generating a single image using a target prompt on an A6000 GPU, reported as the ``Time'' column in Tab.~\ref{table:model_comparison}.
Our method completes the inference in 0.5 seconds, whereas diffusion-based baselines take significantly longer. This speed advantage stems from a shorter progressive inference scheme processing from coarse to fine scales. In contrast, diffusion-based methods require numerous denoising steps within the same dimensional space. 
These results highlight the practical potential of our approach for real-world applications.

Furthermore, we compare our method with CD~\cite{kumari2023multi} and DM~\cite{nam2024dreammatcher}, which achieve the highest quantitative performance among the baselines in subject fidelity and text alignment, respectively. We conduct a user study with 20 participants, collecting a total of 640 responses from 32 comparative questions. Samples are randomly selected from the experimental dataset. Each question presents real subject images, a text prompt, and three generated images from the baseline methods. Users are asked to select the image that best preserves subject identity and prompt fidelity. As shown in Tab.~\ref{table:user_preference}, our method is preferred by (67\%, 79\%) of users over CD and DM, respectively.

\begin{figure*}[t!]
    \includegraphics[width=1.0\textwidth]{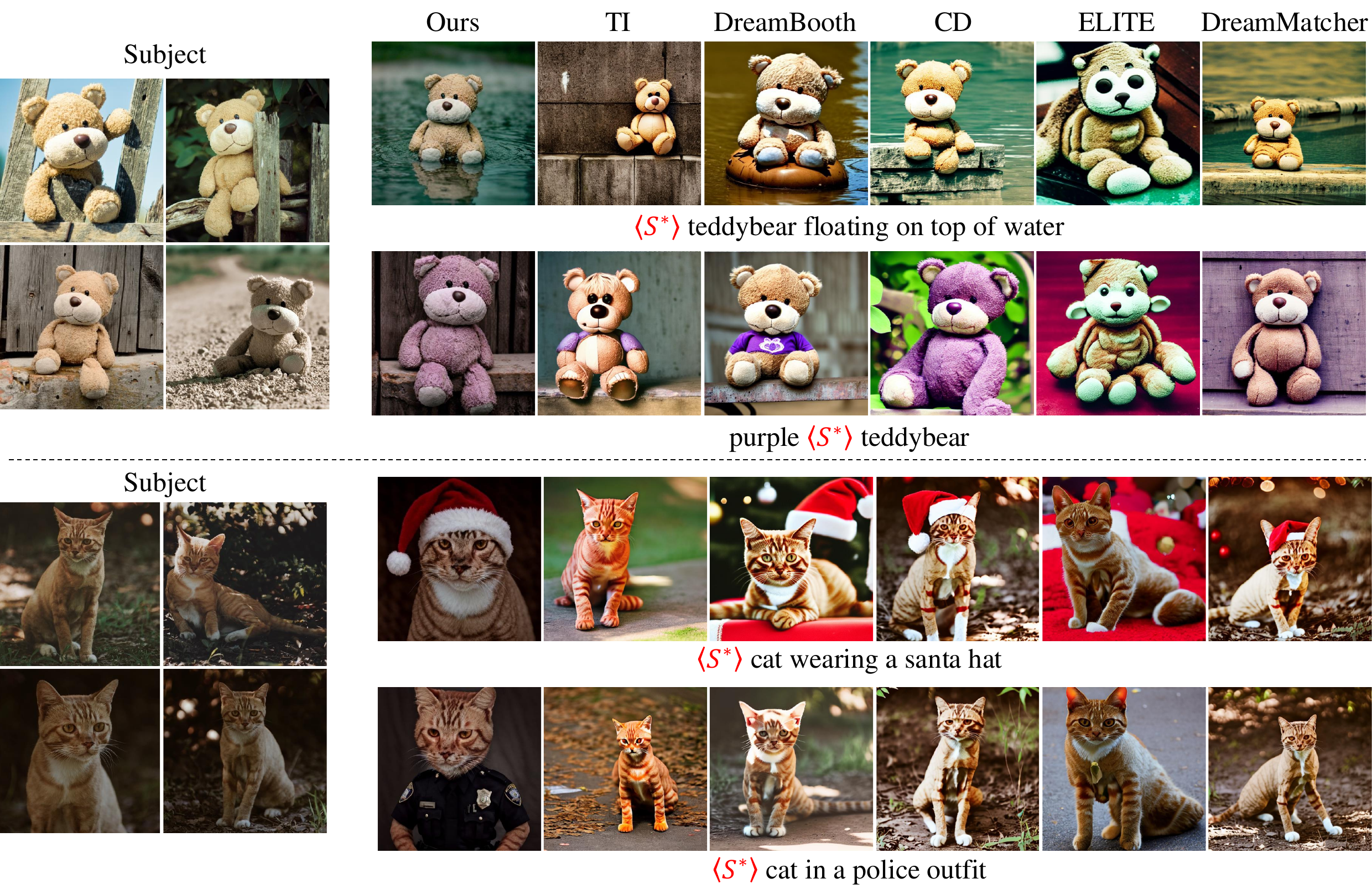}\centering
    \vspace{-0.3cm}
    \caption{Qualitative comparison with diffusion-based baselines.
    }
    \label{figure:main_qualitative_comparison}
    \vspace{-0.4cm}
\end{figure*}

\subsection{Qualitative Comparison}
As shown in Fig.~\ref{figure:main_qualitative_comparison}, our method not only preserves the structure details of the subject image but also aligns closely with the text prompt. For example, in the second row, our approach accurately reconstructs the teddybear's structure, whereas the baseline methods struggle to maintain structural integrity and fails to generate the water details correctly. Furthermore, in the third row, our method successfully generates a cat wearing a Santa hat or a police outfit, while the baseline methods struggle to match the clothing with the cat or fail to generate the appropriate attire.

\subsection{Ablation Study}

\begin{table}[t!]
    \centering
    \resizebox{1.0\linewidth}{!}{%
    \begin{tabular}{l l c c c c c}
        \toprule
         & \textbf{Variant} & \textbf{I\textsubscript{dino}} $\uparrow$ & \textbf{I\textsubscript{clip}} $\uparrow$ & \textbf{T\textsubscript{clip}} $\uparrow$ & \textbf{PRES} $\downarrow$ & \textbf{DIV} $\uparrow$ \\
        \midrule
        \multirow{2}{*}{Sec.~\ref{section:selective_layer_tuning}} & Ours    & 0.786 & \textbf{0.853} & 0.267   & \textbf{0.709} & \textbf{0.272} \\
                                  & w/ Full & \textbf{0.796} & 0.850        & 0.267   & 0.776        & 0.254 \\
        \midrule
        \multirow{2}{*}{Sec.~\ref{section:scale_wise_weighted_tuning}} & Ours    & \textbf{0.786} & \textbf{0.853} & 0.267   & \textbf{0.709} & 0.272 \\
                                  & w/o SWT  & 0.774        & 0.840        & \textbf{0.271}   & 0.792        & \textbf{0.285} \\
        \midrule
        \multirow{2}{*}{Sec.~\ref{section:prior_distillation}} & Ours    & 0.786 & 0.853 & \textbf{0.267}   & \textbf{0.709} & \textbf{0.272} \\
                                  & w/o PD  & \textbf{0.823} & \textbf{0.896} & 0.236   & 0.839        & 0.209 \\
        \bottomrule
    \end{tabular}%
    }
    \vspace{-0.2cm}
    \caption{Quantitative ablation study on proposed strategies.}
    \vspace{-0.4cm}
    \label{table:ablation}
\end{table}

We further analyze the influence of our proposed techniques through comprehensive ablation studies. To this end, we train ablated models on five living subject concepts using 31 prompts, generating eight samples per prompt-subject pair for a total of 1240 images.

\begin{figure}[t!]
\includegraphics[width=1.0\columnwidth]{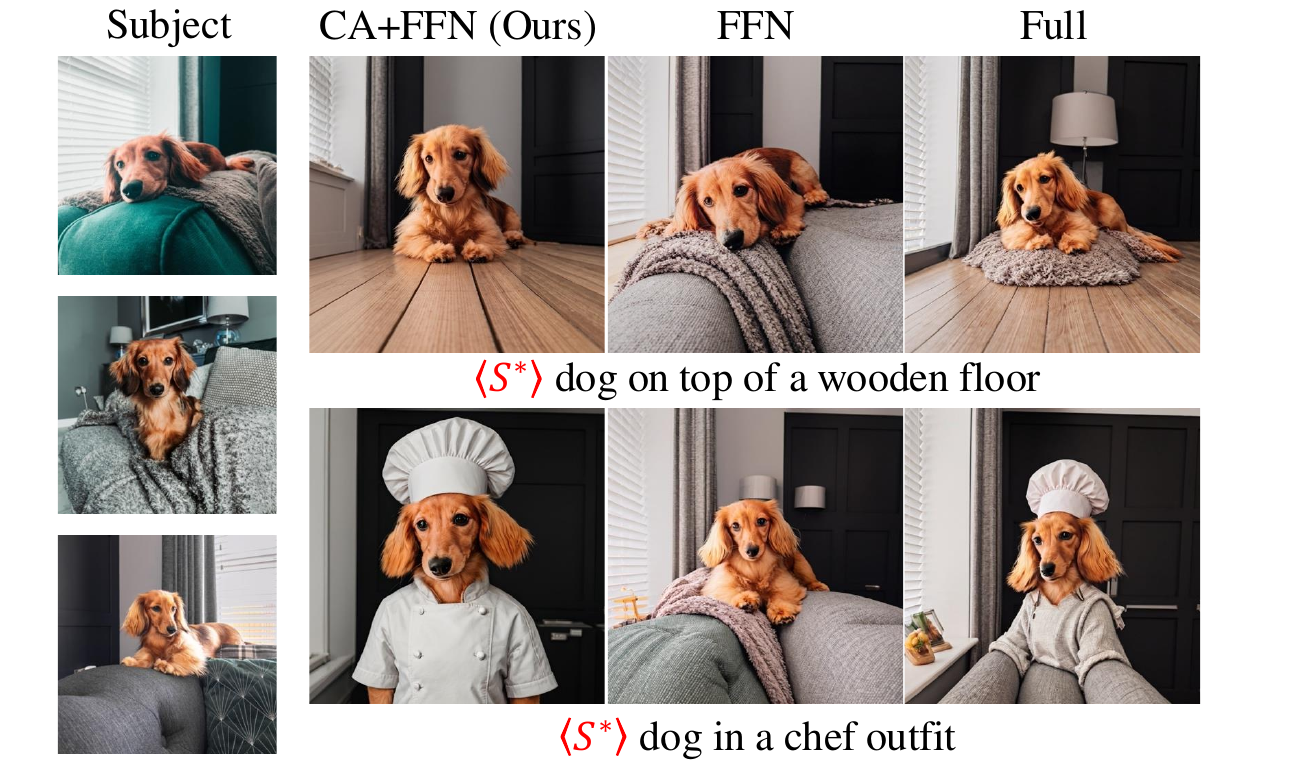}\centering
\vspace{-0.2cm}
    \caption{
    Qualitative comparison of layer selections for tuning.}
    \label{figure:ablation_selective_layer}
    \vspace{-0.5cm}
\end{figure}

\noindent\textbf{Selective layer tuning ablation.}
As shown in the first row of Tab.~\ref{table:ablation}, we compare our model fine-tuned on cross-attention~(CA) and feedforward network~(FFN) layers with full-parameter fine-tuning. While both settings show similar performance, it severely overfit and generate some visual artifacts with fabric, as shown in Fig.~\ref{figure:ablation_selective_layer}. In contrast, our CA and FFN fine-tuning mitigates overfitting, preserves diversity, and reduces computational cost.

\begin{figure}[t!]
\includegraphics[width=0.99\columnwidth]{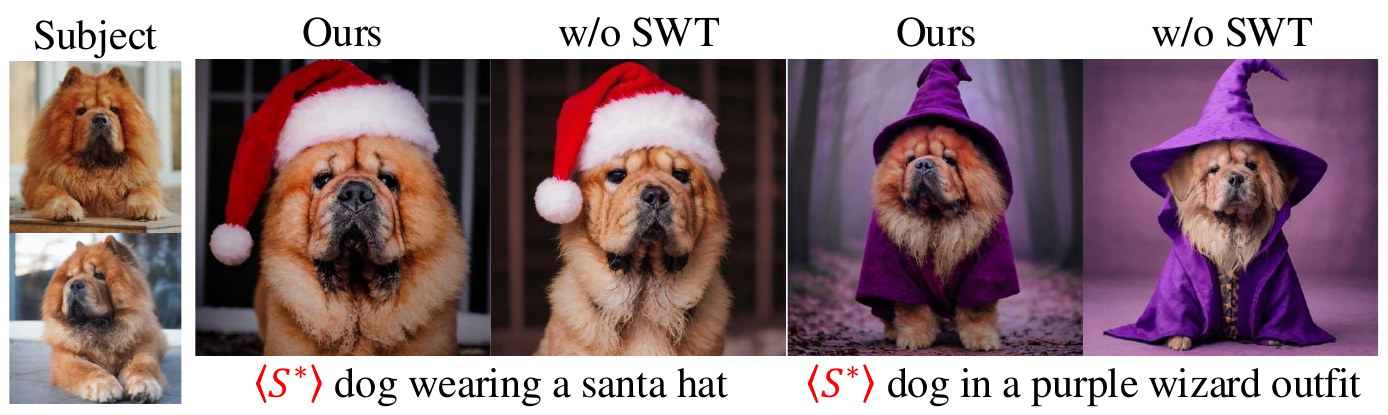}\centering
\vspace{-0.2cm}
    \caption{
    Qualitative comparison with scale-wise weighted tuning.
    }
    \vspace{-0.5cm}
    \label{figure:ablation_scale}
\end{figure}

\noindent\textbf{Scale-wise weighted tuning ablation.}
In our default setting, we use $w_{>8}=0.5$ and set the other weight to $1.0$. To evaluate the impact of weight settings, we ablate using a configuration where all weights are set to $1.0$. As shown in the third row of Tab.~\ref{table:ablation} and Fig.~\ref{figure:ablation_scale}, scale-wise weighted tuning effectively encourages subject fidelity. Although this results in a slight reduction of text alignment, it also preserves prior knowledge, as evidenced by the PRES metrics. This is mainly because scale-wise weighted tuning emphasizes more important coarse-scale token maps inherent in visual autoregressive modeling.

\begin{figure}[t!]
\includegraphics[width=0.95\columnwidth]{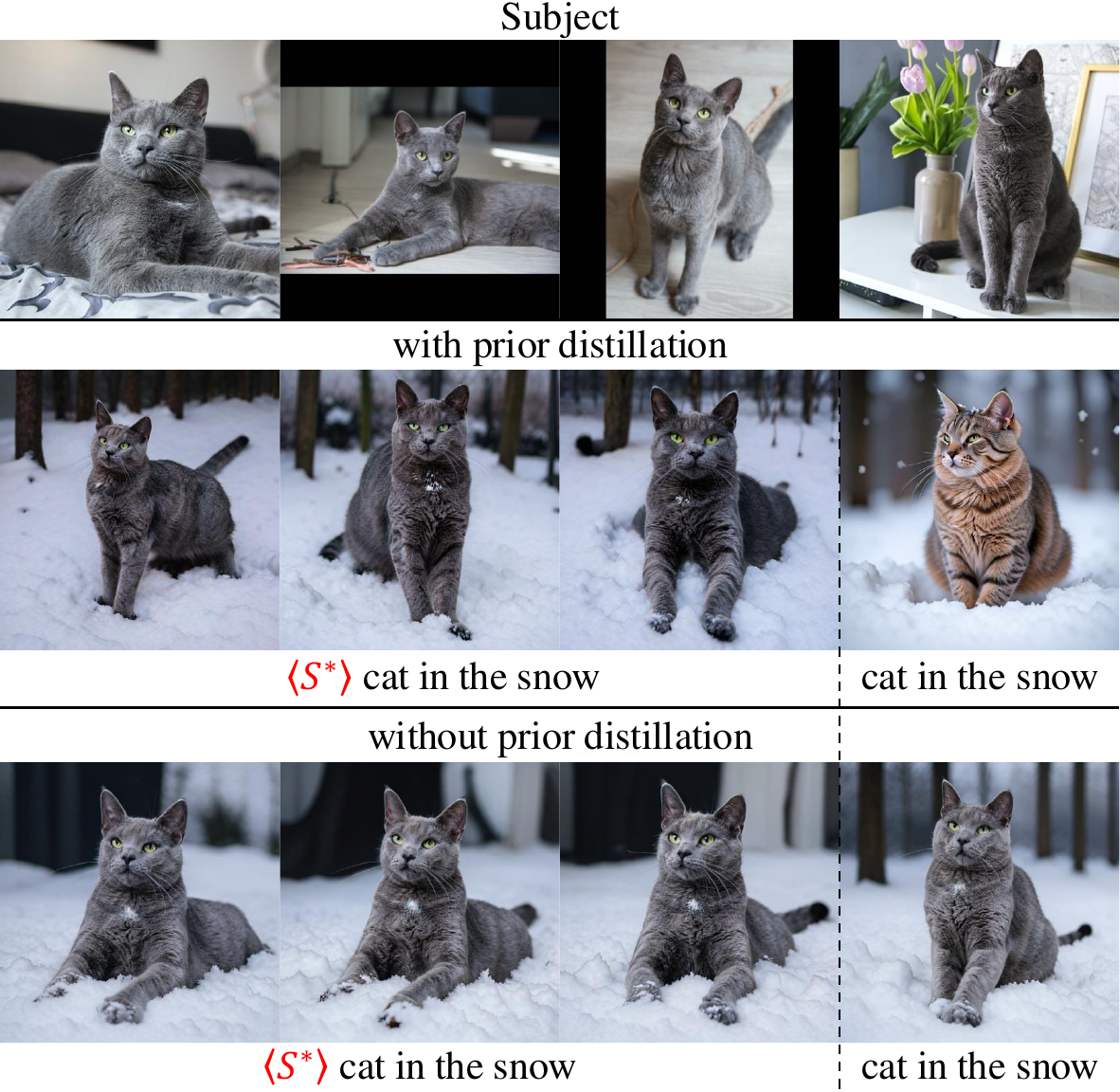}\centering
\vspace{-0.3cm}
    \caption{
    Qualitative comparison with prior distillation.
    }
    \vspace{-0.2cm}
    \label{figure:ablation_prior_distillation}
\end{figure}

\noindent\textbf{Prior distillation ablation.}
We compare our model with and without the proposed prior distillation, which aims to mitigate language drift while preserving original priors. As shown in the third row of Tab.~\ref{table:ablation}, prior distillation effectively counteracts language drift and enhances the model's ability to generate diverse images within prior class. Additionally, DIV metric in Tab.~\ref{table:ablation} indicates that while prior distillation improves diversity, it comes with a slight reduction in subject fidelity, a trend also observed in Dreambooth~\cite{ruiz2023dreambooth}.
This preservation effect is shown in Fig.~\ref{figure:ablation_prior_distillation}, where the model with prior distillation exhibits less overfitting to the reference images and generates cats in more diverse poses.

\subsection{Additional Analysis}
\begin{table}
    \centering
    \resizebox{1.0\linewidth}{!}{%
    \begin{tabular}{lccc|cc}
        \toprule
        \textbf{Model} & \textbf{I\textsubscript{dino}}~$\uparrow$ & \textbf{I\textsubscript{clip}}~$\uparrow$ & \textbf{T\textsubscript{clip}}~$\uparrow$ & \textbf{Time}~$\downarrow$ \\
        \midrule
        SD3.5-md Full     & 0.730 & 0.850 & 0.205 & 15s  \\
        SD3.5-md LoRA & 0.600 & 0.791 & \textbf{0.275} & 15s  \\
        FLUX-LoRA    & 0.573     & 0.793     & 0.273     & 55s  \\
        Ours   & \textbf{0.786} & \textbf{0.853} & 0.267 & \textbf{0.5s} \\
        \bottomrule
    \end{tabular}
    }
    \vspace{-0.3cm}
    \caption{Quantitative comparison with recent diffusion architecture with DreamBooth~\cite{ruiz2023dreambooth}.}
    \label{table:recent_baselines}
    \vspace{-0.4cm}
\end{table}

\begin{figure}[t!]
\includegraphics[width=0.98\columnwidth]{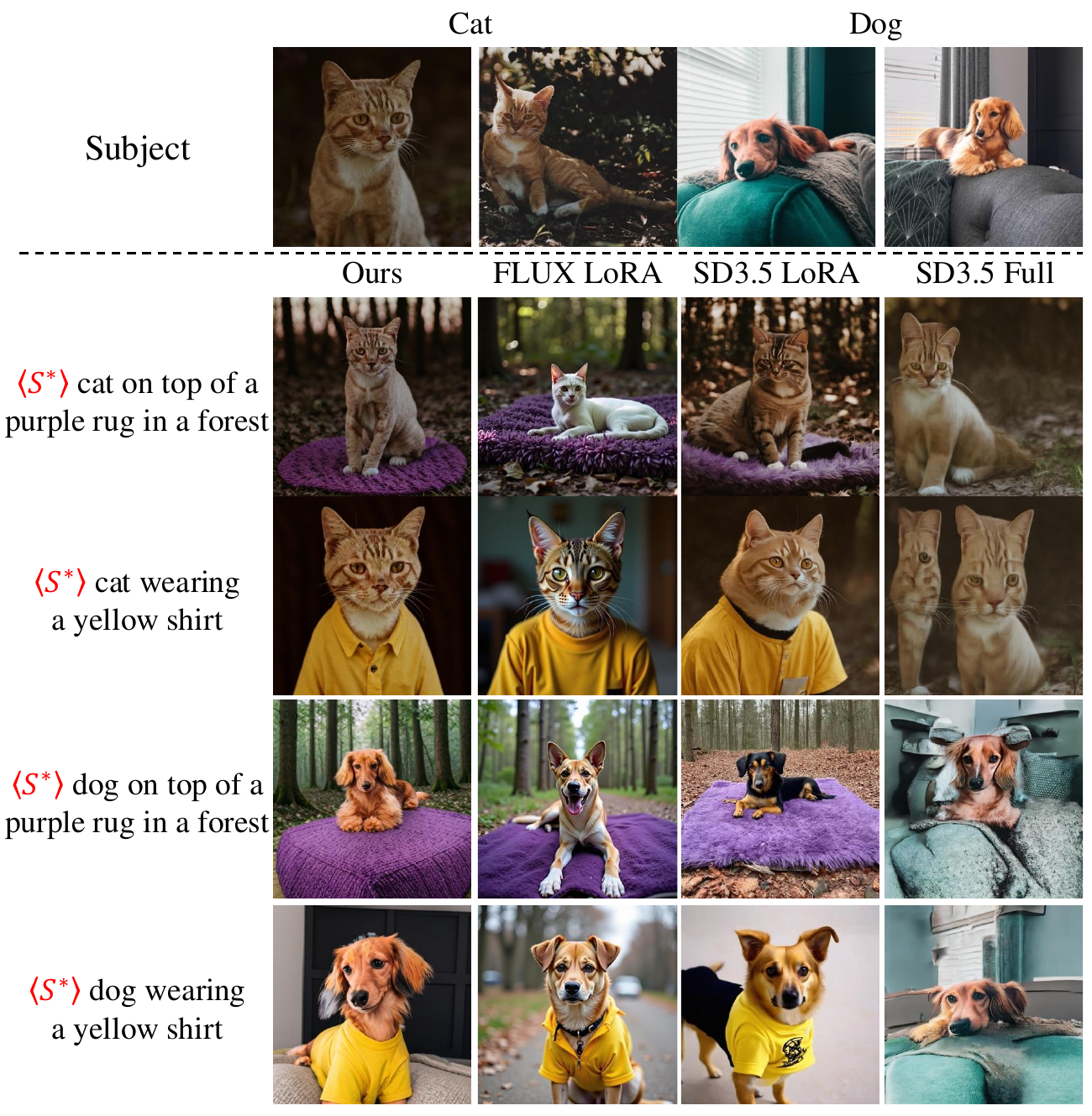}\centering
\vspace{-0.2cm}
    \caption{
    Qualitative comparison with recent diffusion architecture with DreamBooth~\cite{ruiz2023dreambooth}.
    }
    \vspace{-0.4cm}
    \label{figure:recent_compare}
\end{figure}

\noindent\textbf{Comparison with baselines in recent diffusion architectures with DreamBooth.}
Since our base model, Infinity-2B~\cite{han2024infinity}, demonstrates strong generative performance compared to Stable Diffusion~(SD)~1.4, we ensure a fair comparison by also evaluating DreamBooth~\cite{ruiz2023dreambooth} on recently released diffusion architectures, including SD~3.5~\cite{esser2024scaling} and FLUX~\cite{flux2024, labs2025flux1kontextflowmatching}, using LoRA~\cite{hu2022lora} plugins. We train all models for the same number of iterations as our approach and evaluate on 5 living subject concepts.
As shown in Tab.~\ref{table:recent_baselines} and Fig.~\ref{figure:recent_compare}, our generated images achieve the highest subject fidelity among the baselines. While SD~3.5-medium-LoRA exhibits better text alignment, our approach maintains higher subject fidelity. These results indicate that VAR-based approach could surpass recent diffusion-based baselines with substantially faster inference speed.

\noindent\textbf{Optimization Time.}
While our method achieves significantly faster inference than diffusion-based baselines, our optimization time is relatively long, taking approximately 6 hours. For reference, DreamBooth~\cite{ruiz2023dreambooth} applied on SD~1.4 completes optimization in about 30 minutes. However, larger diffusion-based frameworks with comparable performance to Infinity~\cite{han2024infinity}, such as SD~3.5~\cite{esser2024scaling} or FLUX~\cite{flux2024, labs2025flux1kontextflowmatching}, require similar training times to ours. We expect that future work exploring efficient encoder-based tuning methods may further reduce end-to-end duration.

\section{Conclusion}
In this work, we propose the first subject-driven generation approach using a large-scale visual autoregressive~(VAR) model. We introduce selective layer tuning, scale-wise weighted tuning, and prior distillation to address language drift, computational overhead, and reduced diversity.
Experiments validate that our method outperforms diffusion-based baselines in subject fidelity, text alignment, and inference speed.

\section*{Acknowledgements}
\vspace{-0.2cm}
This work was supported in part by MSIT/IITP (No. RS-2022-II220680, 2020-0-01821, RS-2019-II190421, RS-2024-00459618, RS-2024-00360227, RS-2024-00437633), MSIT/NRF (No. RS-2024-00357729), KNPA/KIPoT (No. RS-2025-25393280), and SEMES-SKKU collaboration funded by SEMES.

{
    \small
    \bibliographystyle{ieeenat_fullname}
    \bibliography{main}
}

\clearpage
\appendix
\twocolumn[{
    \begin{center}
        \vspace{0.5cm}
        {\Large \bf Supplementary Material \par}
        \vspace{0.3cm}
    \end{center}
}]
\section{Additional Ablations}

\subsection{Scale-wise Weighted Tuning for Style Personalization}

\begin{figure}[t!]
    \includegraphics[width=1.0\columnwidth]{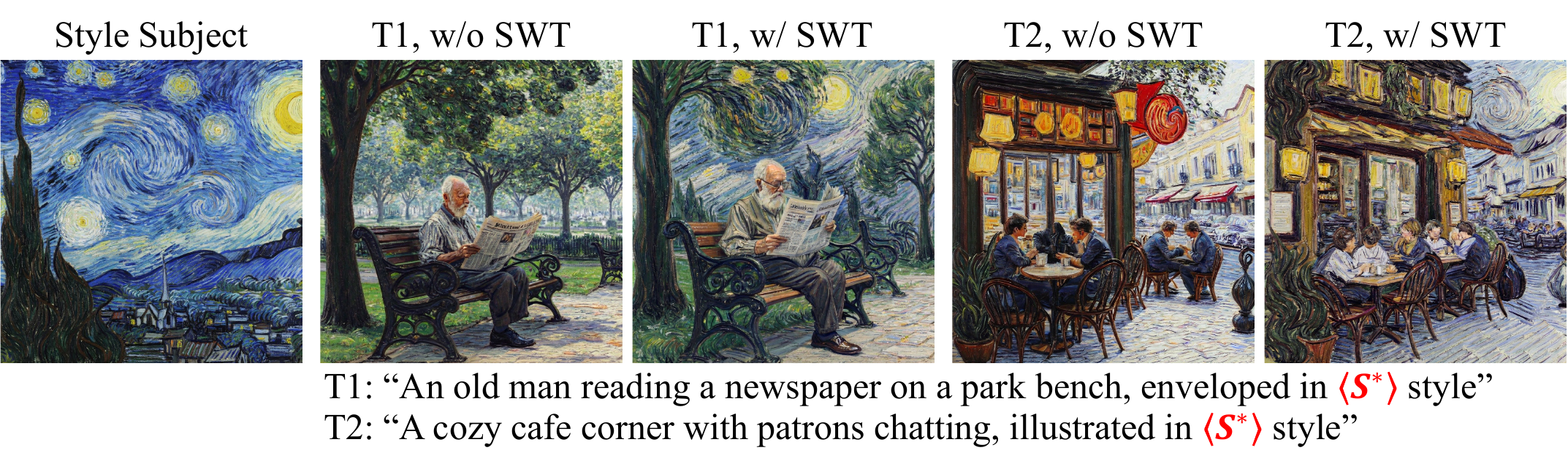}
    \centering
    \vspace{-0.6cm}
    \caption{Qualitative comparison demonstrating the effectiveness of Scale-wise Weighted Tuning (SWT) for style personalization.
    }
    \label{figure:ablation_style}
\end{figure}

Unlike diffusion models, where later timesteps capture fine, high-frequency details, our analysis indicates that later scales of large-scale pretrained VAR models contribute minimally to output variation. Consequently, our proposed Scale-wise Weighted Tuning (SWT) strategy de-emphasizes these scales, improving robustness in style personalization tasks, particularly for capturing high-frequency details.

To validate SWT effectiveness, we fine-tune on eight distinct style concepts from DreamBench++~\cite{pengdreambench++}, each with nine text prompts, for 100 iterations per style. We generate eight outputs per pair and present representative qualitative comparisons in Fig.~\ref{figure:ablation_style}. As shown in Fig.~\ref{figure:ablation_style}, SWT achieves clearer stylistic expressions, such as distinct swirls and brushstrokes.

\subsection{Prior Distillation vs. Prior Preservation Loss}

We further compare our Prior Distillation (PD) method against the Prior Preservation Loss (PPL) approach~\cite{ruiz2023dreambooth}. For integrating PPL into VAR~\cite{tian2025visual}, we replace the original MSE objective with cross-entropy objective, using 100 class-specific images generated with the Infinity-2B checkpoint.

\begin{figure}[t!]
\centering
\includegraphics[width=1.0\columnwidth]{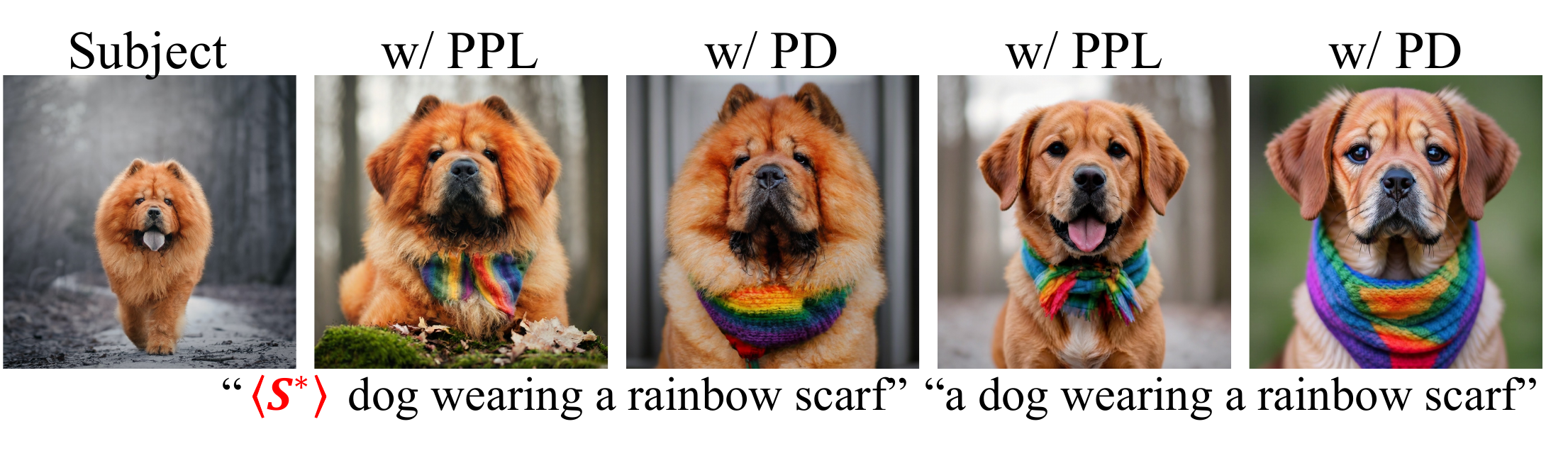}
\vspace{-0.6cm}
\caption{Qualitative comparison between our Prior Distillation (PD) and Prior Preservation Loss (PPL).}
\label{figure:ppl}
\end{figure}

Quantitative results in Tab.~\ref{tab:ppl_comparison} and qualitative results in Fig.~\ref{figure:ppl} demonstrate comparable performance between PD and PPL across various metrics. Notably, PD eliminates the requirement for same-class dataset collection, substantially reducing the preparation time and computational overhead.

\begin{table}
\centering
\setlength\tabcolsep{3pt}
\begin{tabular}{cccccc}
\toprule
Variant & \textbf{I\textsubscript{dino}} $\uparrow$ & \textbf{I\textsubscript{clip}} $\uparrow$ & \textbf{T\textsubscript{clip}} $\uparrow$ & \textbf{PRES} $\downarrow$ & \textbf{DIV} $\uparrow$ \\ 
\midrule
Ours w/ PD & 0.786 & 0.853 & 0.267 & 0.709 & 0.272 \\
Ours w/ PPL & 0.653 & 0.822 & 0.265 & 0.608 & 0.378 \\
\bottomrule
\end{tabular}
\caption{Quantitative comparison between our Prior Distillation (PD) and Prior Preservation Loss (PPL).}
\label{tab:ppl_comparison}
\end{table}

\subsection{Selective Layer Tuning with LoRA}

We evaluate Selective Layer Tuning (SLT) combined with Prior Distillation (PD) against the popular LoRA method~\cite{hu2022lora}. Specifically, we apply LoRA across all VAR layers at multiple ranks (4 and 16) without employing SLT or PD, to isolate their contributions. As shown in Tab.~\ref{table:lora_comp}, LoRA without SLT or PD results in unstable fine-tuning, causing visual artifacts or failing to accurately capture subject concepts.

\begin{figure}[t!]
\centering
\includegraphics[width=1.0\columnwidth]{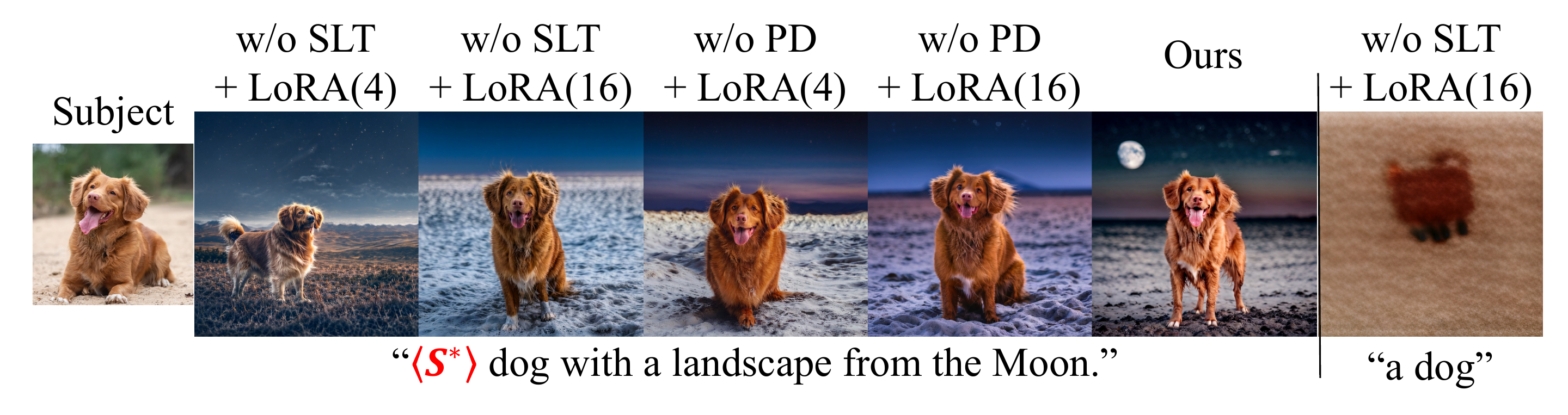}
\vspace{-0.6cm}
\caption{Qualitative comparison of LoRA adaptation versus our SLT + PD approach.}
\label{figure:lora}
\end{figure}

The qualitative results in Fig.~\ref{figure:lora} further confirm that our combined approach (SLT + PD) stabilizes personalization and preserves subject identity, validating their critical role in achieving robust VAR personalization.

\begin{table}
\centering
\footnotesize
\setlength\tabcolsep{3pt}
\begin{tabular}{c|c|ccccc}
\toprule
\textbf{Base} & \textbf{Adapter} & \textbf{I\textsubscript{dino}} $\uparrow$ & \textbf{I\textsubscript{clip}} $\uparrow$ & \textbf{T\textsubscript{clip}} $\uparrow$ & \textbf{PRES} $\downarrow$ & \textbf{DIV} $\uparrow$ \\
\midrule
\multirow{2}{*}{w/o SLT} & +~LoRA(4) & 0.653 & 0.822 & 0.265 & 0.608 & 0.378 \\
                         & +~LoRA(16) & 0.770 & 0.852 & 0.264 & 0.383 & 0.351 \\
\midrule
\multirow{2}{*}{w/o PD}  & +~LoRA(4) & 0.753 & 0.852 & 0.265 & 0.676 & 0.376 \\
                         & +~LoRA(16) & 0.775 & 0.854 & 0.265 & 0.781 & 0.373 \\
\midrule
\multicolumn{2}{c|}{Ours~(w/ SLT, PD)} & 0.786 & 0.853 & 0.267 & 0.709 & 0.272 \\
\bottomrule
\end{tabular}
\caption{Quantitative comparison with LoRA.}
\label{table:lora_comp}
\end{table}

\section{User Study Details}

\begin{figure}[t!]
\centering
\includegraphics[width=0.99\columnwidth]{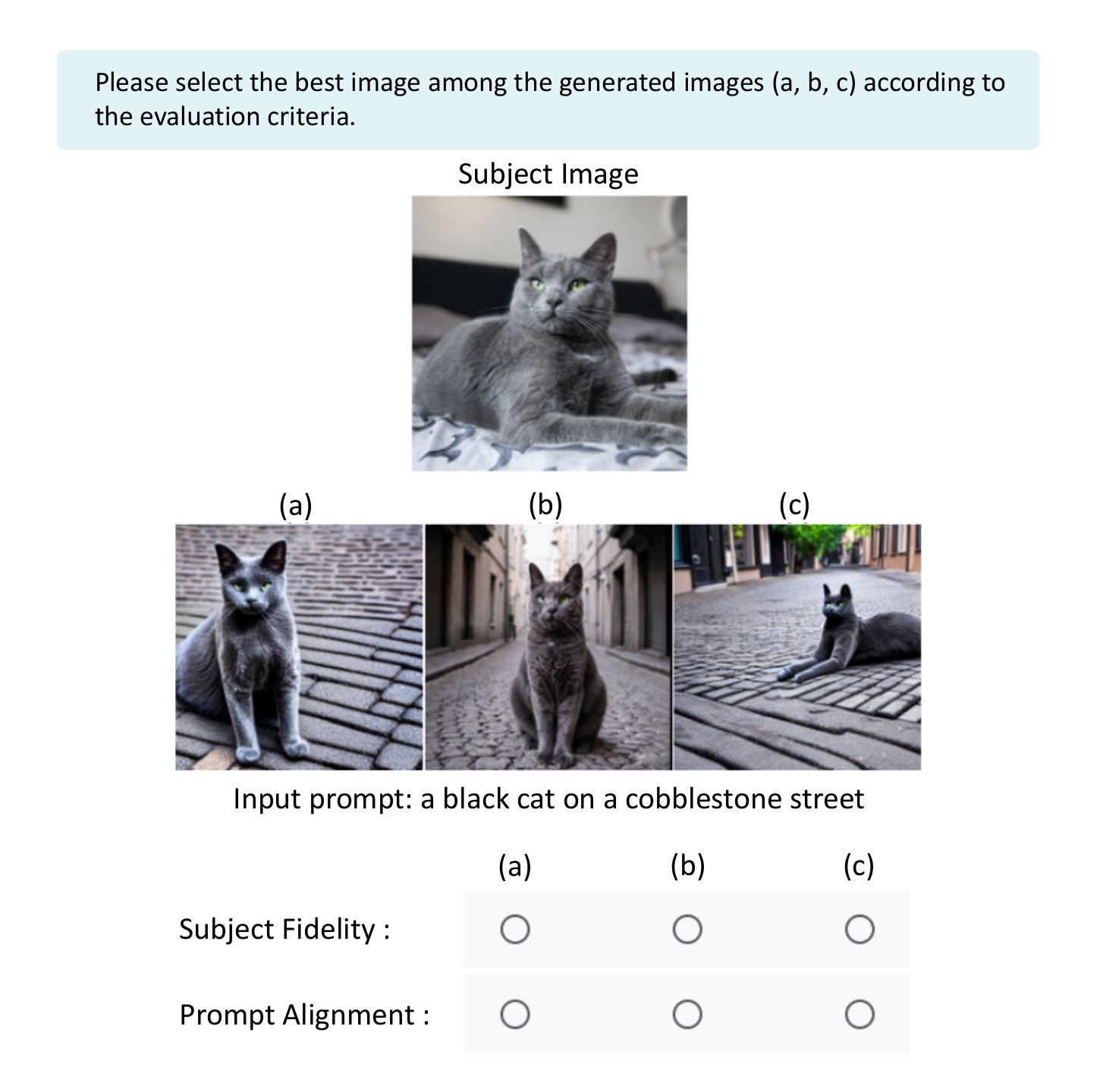}
\vspace{-0.3cm}
\caption{User study interface for evaluating personalization quality.}
\label{figure:user_study}
\end{figure}

We conduct a user study with 20 participants to evaluate personalization effectiveness. Participants compare outputs from our method against baselines, focusing on subject fidelity and prompt alignment, following the evaluation protocol from DreamBooth~\cite{ruiz2023dreambooth}. The user interface is shown in Fig.~\ref{figure:user_study}.

\section{Additional Qualitative Comparisons}

\subsection{Comparison with FLUX-based Method}
\begin{figure}[t!]
\centering
\includegraphics[width=0.99\columnwidth]{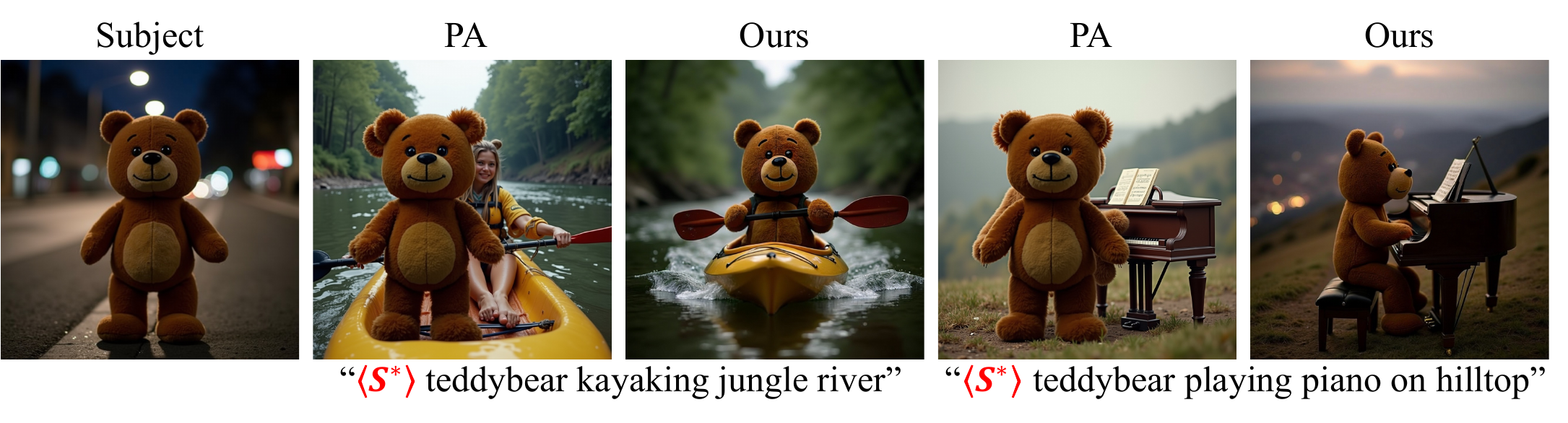}
\vspace{-0.3cm}
\caption{Qualitative comparison to Personalize Anything (PA)~\cite{feng2025personalize}.}
\label{figure:pa}
\end{figure}

We further compare our VAR-based approach with Personalize Anything (PA)~\cite{feng2025personalize}, which is based on the FLUX framework~\cite{labs2025flux1kontextflowmatching, flux2024}. As illustrated in Fig.~\ref{figure:pa}, PA~\cite{feng2025personalize} struggles to generate dynamic poses from the subject image, whereas our approach successfully synthesizes diverse and dynamic poses, highlighting the efficacy of our VAR-based personalization.

\subsection{Comparison with Diffusion-based Methods}
\begin{figure*}[t!]
\centering
\includegraphics[width=0.99\textwidth]{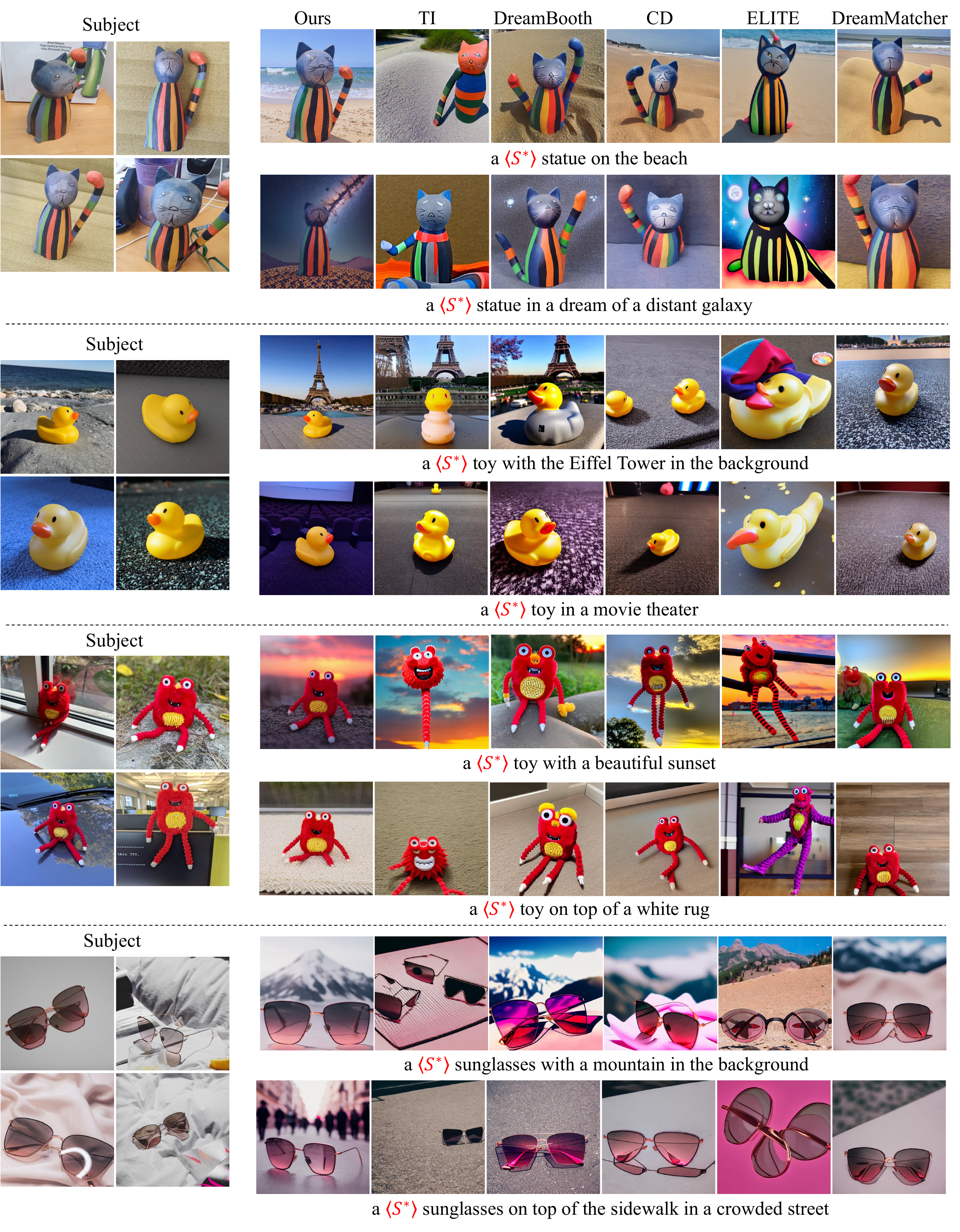}
\caption{Extended qualitative comparisons to diffusion-based baselines.}
\label{figure:sup_qual}
\end{figure*}

Additional qualitative comparisons against diffusion-based baselines are provided in Fig.~\ref{figure:sup_qual}. These examples clearly demonstrate our method's ability to better preserve subject identity, accurately capturing crucial attributes such as color and shape. Furthermore, our method achieves notably improved alignment with the provided prompts.

\end{document}